\documentclass[11pt,letterpaper]{article}
\usepackage{eacl2017}
\usepackage{times}
\usepackage{latexsym}
\usepackage{url}
\usepackage{graphicx}
\usepackage{booktabs}
\usepackage{adjustbox}
\usepackage{svg}
\usepackage{tikz}
\usepackage{gb4e} 
\noautomath 

\eaclfinalcopy 

\usepackage[normalem]{ulem}

\title{Story Cloze Ending Selection Baselines and Data Examination}

\author{Todor Mihaylov \\
Research Training Group AIPHES\\
Institute for Computational Linguistics\\
Heidelberg University\\
  {\tt \small{mihaylov@cl.uni-heidelberg.de}} \\
\\\And
  Anette Frank \\
  Research Training Group AIPHES\\
  Institute for Computational Linguistics\\
  Heidelberg University\\
  {\tt \small{frank@cl.uni-heidelberg.de}}\\} 
  
\begin{document}
\maketitle
\begin{abstract}
This paper describes two supervised baseline systems for the Story Cloze Test Shared Task \cite{Mostafazadeh2016AStories}. We first build a classifier using features based on word embeddings and semantic similarity computation. We further implement a neural LSTM system with different 
encoding strategies
that try to model the relation between the story and the provided endings. Our 
experiments show that a model using representation features based on average word embedding vectors over the given story words and the candidate ending sentences words, joint with  similarity features between the story and candidate ending representations performed better than the neural models. Our best model 
achieves an accuracy of 72.42, ranking 3rd in the official evaluation. 
\end{abstract}

\section{Introduction}
\label{sec:intro}
Understanding common sense stories is an easy task for humans but represents a challenge for machines. A recent common sense story under\-stan\-ding task is the 'Story Cloze Test' \cite{Mostafazadeh2016AStories}, where a human or an AI system has to read a given four-sentence story and select the proper ending out of two proposed endings. While the majority class baseline performance on the given test set yields an accuracy of 51.3\%, human performance achieves 100\%. This makes the task a good challenge for an AI system. The Story Cloze Test task is proposed as a Shared Task for LSDSem 2017\footnote{Workshop on Linking Models of Lexical, Sentential and Discourse-level Semantics 2017}. 17 teams registered for the Shared Task and 10 teams submitted their results\footnote{https://competitions.codalab.org/competitions/15333 - Story Cloze Test at CodaLab}.
Our contribution is that we set a new baseline for the task, showing that a simple linear model based on distributed representations and semantic similarity features achieves state-of-the-art results. We also evaluate the ability of different embedding models to represent common knowledge required for this task. We present an LSTM-based classifier with different representation encodings that tries to model the relation between the story and alternative endings and argue about its inability to do so. 

\section{Task description and data construction}\label{sec:data}

The Story Cloze Test is a natural language understanding task that consists in selecting the right ending for a given short story. 
The evaluation data consists of a {\em Dev set} and a {\em Test set}, each containing samples of four sentences of a story, followed by two alternative sentences, from which the system has to select the proper story ending.
An example of such a story is presented in Table \ref{table:story-cloze-sample}.

\begin{table*}[t]
\centering
\begin{adjustbox}{max width=\textwidth}
\begin{tabular}{@{}p{0.47\textwidth}p{0.225\textwidth}p{0.225\textwidth}@{}}
\textbf{Story context}  & \textbf{Good Ending}  & \textbf{Bad ending}                   \\\hline
Mary and Emma drove to the beach.  They decided to swim in the ocean.  Mary turned to talk to Emma.  Emma said to watch out for the waves. 
 & A big wave knocked Mary down. & The ocean was a calm as a bathtub.\\\hline
\end{tabular}
\end{adjustbox}
\caption{Example of a given story with a bad and a good ending.}
\label{table:story-cloze-sample}
\end{table*}

The instances in the  \textit{Dev} and \textit{Test} gold data sets (1871 instances each) were crowd-sourced together with the related ROC Stories corpus \cite{Mostafazadeh2016AStories}. The ROC stories consists of around 100,000 crowd-sourced short five sentence 
stories ranging over various topics. These stories do not feature a wrong ending, but with appropriate extensions they can be deployed as training data for the Story Cloze task.


\paragraph{Task modeling.}
We approach the task as a supervised classification problem.
For every classification instance \textit{(Story, Ending1, Ending2}) we predict one of the labels in Label=\textit{\{Good,Bad\}}.

\paragraph{Obtaining a small training set from \textit{Dev} set.}
We construct a (small)  training data set from the \textit{Dev} set by splitting it randomly into a \textit{Dev-Train} and a \textit{Dev-Dev} set containing 90\% and 10\% of the original \textit{Dev} set. From each instance in \textit{Dev-Train}
we generate 2 instances by swapping \textit{Ending1} and \textit{Ending2} and inversing the class label.

\paragraph{Generating training data from ROC stories.} 
We also make use of the 
ROC Stories corpus in order to generate a large training data set. 
We experiment with three methods:

{\bf (i.) Random endings.} For each story 
we employ the first 4 sentences as the story context. We use the original ending as \textit{Good} ending and define a \textit{Bad} ending by randomly choosing some ending from an alternative story in the corpus.
From each story with one {\em Good} ending we generate 10 \textit{Bad} examples by selecting 10 random endings.

{\bf (ii.) Coherent stories and endings with common participants and noun arguments.} Given that some random story endings are too clearly unconnected to the story, 
here we aim to select {\em Bad} candidate endings that are coherent with the story, yet still distinct from a {\em Good} ending. 
To this end, for each story in the ROC Stories corpus, we obtain the lemmas of all pronouns (tokens with part of speech tag starting with `PR`) and lemmas of all nouns (tokens with part of speech tag starting with `NN`) and select the top 10 endings from other stories that share most of these
features as \textit{Bad} endings. 

{\bf (iii.) Random coherent stories and endings.} We also modify (ii.) so that we select the nearest 500 endings to the story context and select 10 randomly. 

\section{A Baseline Method}
\label{sec:method-baseline}
For tackling the problem of right story ending selection we follow a feature-based classification approach that was previously applied to bi-clausal classification tasks in \cite{mihaylovfrank:2016,SemEval2016:task3:SemanticZ}. 
It uses features based on word embeddings to represent the clauses and semantic similarity measured between these representations for the clauses. Here, we adopt this approach to model parts of the story and the candidate endings. 
For the given \textit{Story} and the given candidate \textit{Endings} we extract features based on word embeddings. An advantage of this approach is that it is fast for training and that it only requires pre-trained word embeddings as an input.

\subsection{Features}
\label{sec:method-baseline:features}

In our models we only deploy features based on word embedding vectors. 
We are using two types of features: (i) {\bf representation features} that model the semantics of parts of the story using word embedding vectors, and (ii) {\bf similarity scores} that capture specific properties of the relation holding between the story and its candidate endings. For computing similarity between the embedding representations of the story components, we employ cosine similarity.

The different feature types are described below.


\paragraph{(i) Embedding representations for \textit{Story} and \textit{Ending}.}
For each \textit{Story} (sentences 1 to 4) and story endings \textit{Ending1} and \textit{Ending2} we construct a centroid vector 
from the embedding vectors $\vec{w_i}$ of all words $w_i$ in their respective surface yield.



\paragraph{(ii.) Story to Ending Semantic Vector Similarities.} We calculate various similarity features on the basis of the centroid word vectors for 
all or selected sentences in the given \textit{Story} and the \textit{Ending1} and \textit{Ending2} sentences, as well as on parts of the these sentences:

\subparagraph{Story to Ending similarity.} We assume that a given \textit{Story} and its \textit{Good Ending} are connected by a specific semantic relation or some piece of common sense knowledge. Their representation vectors should thus stand in a specific similarity relation to each other. We use their cosine similarity as a feature. Similarity between the story sentences and a candidate ending has already been proposed as a baseline by \newcite{mostafazadeh-EtAl:2016:RepEval} but it does not perform well as a single feature.  

\subparagraph{Maximized similarity.} This measure ranks each word in the \textit{Story} according to its similarity with the centroid vector of \textit{Ending}, and we compute the average similarity for the top-ranked $N$ words. We chose the similarity scores of the top 1,2,3 and 5 words as features. 
Our assumption is that the average similarity between the {\em Story} representation and the top $N$ most similar words in the \textit{Ending} might characterize the proper ending as the {\em Good} ending. 
We also extract {\bf maximized aligned similarity}. For each word in \textit{Story}, we choose the most similar word from the yield of \textit{Ending} and take the average of all best word pair similarities, as suggested in \newcite{tran-EtAl:2015:SemEval}.
\subparagraph{Part of speech (POS) based word vector similarities.}
For each sentence in the given four sentence story and the candidate endings we performed part of speech tagging using the Stanford CoreNLP \cite{Manning2014CoreNlp} parser,
and computed similarities between centroid vectors of words with a specific tag from \textit{Story} and the centroid vector of \textit{Ending}.
Extracted features for POS similarities include symmetric and asymmetric combinations: for example we calculate the similarity between \textit{Nouns} from \textit{Story} with \textit{Nouns} from \textit{Ending} and similarity between \textit{Nouns} from \textit{Story} with \textit{Verbs} from \textit{Ending} and vice versa.

The assumption is that embeddings for some parts of speech between \textit{Story} and \textit{Ending} might be closer to those of other parts of speech for the {\em Good} ending 
of a given story. 

\subsection{Classifier settings}
\label{sec:method-baseline:logreg}
For our feature-based approach we concatenate the extracted representation and similarity features 
in a feature vector, scale their values to the 0 to 1 range, and feed the vectors to a classifier. We train and evaluate a L2-regularized Logistic Regression classifier with the LIBLINEAR \cite{Fan2008Liblinear} solver as implemented in {\em scikit-learn} \cite{scikit-learn}.

For each separate experiment we tune the regularization parameter C with 5 fold cross-validation on the \textit{Dev} set and then train a new model on the entire \textit{Dev} set in order to evaluate on the \textit{Test} set.

\section{Neural LSTM Baseline Method}
\label{sec:neural-method}
We compare our feature-based linear classifier baseline to a neural approach. Our goal is to explore a simple neural method and to investigate how well it performs with the given small dataset.
We implement a Long Short-Term Memory (LSTM) \cite{Hochreiter:1997:LSM:1246443.1246450} recurrent neural network model.

\subsection{Representations}
\label{sec:neural-method:representation}

We are using the raw LSTM output of the encoder. 
We also experiment with an encoder with attention to model the relation between a story and a candidate ending, following \cite{Rocktaschel2015}. 

\paragraph{(i) Raw LSTM representations.}
For each given instance \textit{(Story, Ending1, Ending2)} we first encode the \textit{Story} token word vector representations using a recurrent neural network (RNN) with long short-term memory (LSTM) units. We use the last output \begin{math}\textbf{h}_{L}^s\end{math} and \begin{math}\textbf{c}_{L}^s\end{math} states of the \textit{Story} to initialize the first LSTM cells for the respective encodings \begin{math}\textbf{e}_{1}\end{math} and \begin{math}\textbf{e}_{2}\end{math} of {\em Ending1} and {\em Ending2}, where \begin{math}\textbf{L}\end{math} is the token length of the \textit{Story} sequence. 

We build the final representation \begin{math}\textbf{o}_{se1e2}\end{math} by concatenating the \begin{math}\textbf{e}_{1}\end{math} and \begin{math}\textbf{e}_{2}\end{math} representations. 
Finally, for classification we use a softmax layer over the output \begin{math}\textbf{o}_{se1e2}\end{math} by mapping it into the target space of the two classes (Good, Bad) using a parameter matrix \begin{math}\textbf{M}_{o}\end{math} and bias \begin{math}\textbf{b}_{o}\end{math} as given in (Eq.\ref{eq:softmax:lstm}). We train using the cross-entropy loss.
\begin{equation} \label{eq:softmax:lstm}
label_{prob} = softmax({W^o}{o}_{se1e2} + {b}_{o})
\end{equation}


\paragraph{(ii) Attention-based representations}
We also model the relation \begin{math}h^{*}\end{math} between the \textit{Story} and each of the \textit{Endings} using the attention-weighted representation \begin{math}\textbf{r}\end{math} between the last token output \begin{math}\textbf{h}_{N}^e\end{math} of the \textit{Ending} representation and each of the token representations \begin{math}[\textbf{h}_{1}^s..\textbf{h}_{L}^s]\end{math} of the \textit{Story}, strictly following the attention definition by \newcite{Rocktaschel2015}. The final representation for each ending is presented by Eq.\ref{eq:att:endings}, where \begin{math}\textbf{W}^p\end{math} and \begin{math}\textbf{W}^x\end{math} are trained projection matrices.
 
\begin{equation} \label{eq:att:endings}
h^{*} = tanh({W^p}r+{W^x}h^e_N)
\end{equation}

We then present the output representation \begin{math}\textbf{o}_{se1e2}\end{math} as a concatenation of the encoded \textit{Ending1} and \textit{Ending2} representations \begin{math}h^{*}_{e1}\end{math} and \begin{math}h^{*}_{e2}\end{math} and use Eq.\ref{eq:softmax:lstm} to obtain the output label likelihood.

\paragraph{(iii) Combined raw LSTM output and attention representation} We also perform experiments with combined LSTM outputs and representations. In this setting we present the output \begin{math}\textbf{o}_{se1e2}\end{math} as presented in Eq.\ref{eq:repr:concat}: 
\begin{equation} \label{eq:repr:concat}
o_{se1e2}= concat(e_1, h^{*}_{e1}, e_2, h^{*}_{e2})
\end{equation}
\subsection{Model Parameters and Tuning}
\label{sec:neural-method:parameters}
  We perform experiments with configurations of the model using grid search on the batch size (50, 100, 200, 300, 400, 500) and LSTM output size (128, 256, 384, 512), by training a simple model with raw LSTM encoding on \textit{Dev-Train} and evaluating on the \textit{Dev-Dev}. For each configuration we train 5 models and take the parameters of the best. The best result on the \textit{Dev-Dev} set is achieved for LSTM with output size 384 and batch Size 500 after 7 epochs and achieves accuracy of 72.10 on the official \textit{Test}. For learning rate optimization we use Adam optimizer \cite{Kingma2015} with initial learning rate 0.001.
  \paragraph{Parameter initialization.} We initialize the LSTM weights with Xavier initialization \cite{Glorot2010} and bias with a constant zero vector.
 
\section{Experiments and Results}
\label{sec:experiments}

\paragraph{Overall results.}
In Table \ref{table:baseline-results} we compare our best systems to existing baselines, Shared Task participant systems\footnote{https://competitions.codalab.org/competitions/15333 - The Story Cloze Test Shared Task home page} and human performance. \textit{Our features baseline} system is our best feature-based system using embeddings and \textit{word2vec} trained on \textit{Dev} and tuned with cross-validation. \textit{Our neural system} employs raw LSTM encodings as described in Section \ref{sec:neural-method:representation}(i) and it is trained on the \textit{Dev-Dev} dataset which consists of 90\% of the \textit{Dev} dataset selected randomly and tuned on the rest of \textit{Dev}.
The best result in the task is achieved by \newcite{Schwartz2017-story-cloze} \textit{(msap)} who employ stylistic features combined with RNN representations. We have no information about \textit{cogcomp} and \textit{ukp}.

\begin{table}[]
\centering
\begin{tabular}{ll}
\textbf{System} & \textbf{Accuracy} \\\hline
Human         & 100.00   \\\hline
msap & 75.20   \\
cogcomp & 74.39   \\
\textbf{Our features baseline}     & \textbf{72.42}   \\
\textbf{Our neural system}     & \textbf{72.10}   \\
ukp & 71.67    \\
%
%

DSSM          & 58.50    \\
Skip-thoughts sim & 55.20    \\
Word2Vec sim  & 53.90    \\
Majority baseline      & 51.30    \\\hline
\end{tabular}
\caption{Comparison of our models to  shared task participants' results and other baselines. \textit{Word2Vec sim}, \textit{Skip-thoughts sim} and \textit{DSSM} are described in \cite{mostafazadeh-EtAl:2016:RepEval}.}
\label{table:baseline-results}
\end{table}


\paragraph{Model variations and experiments.}



The Story Cloze Test is a story understanding problem. However, the given stories are very short and they require background knowledge about 
relations between the given entities, entity types and events defining the story and their endings, as well as relations between these events. 
We first train our feature-based model with alternative embedding representations  in order to select the best source of knowledge for further experiments.

\begin{table*}[h]
\centering
\begin{adjustbox}{max width=\textwidth}
\begin{tabular}{lccccccc}
\textbf{Model} & \textbf{All} & \textbf{All wo POS sim} & \textbf{All wo MaxSim} & \textbf{All wo Sim} & \textbf{WE S, E1, E2+Sim} & \textbf{WE  E1, E2} & \textbf{Sims only} \\\hline
Word2Vec GN 100B 300d & \textbf{72.42} & 71.41 & 71.94 & 72.10 & 71.51 & 70.71 & 58.15\\
Concepnet 300d & 72.05 & \textbf{72.05} & \textbf{72.05} & 72.05 & \textbf{71.83} & \textbf{71.67}  & 61.67\\
Glove  840B 300d & 71.41 & 71.09 & 71.89 & \textbf{72.26} & 70.82 & 70.71  & 60.28\\
Glove  6B 200d & 69.43 & 69.75 & 68.31 & 69.64 & 68.04 & 68.68  & \textbf{62.37}\\
Glove  6B 300d & 68.84 & 69.32 & 69.21 & 69.05 & 68.79 & 68.89  & 61.19\\
Glove  6B 100d & 68.84 & 68.09 & 67.93 & 68.41 & 67.66 & 67.56  & 60.82\\
Glove  6B 50d & 64.89 & 66.01 & 64.19 & 64.67 & 64.78 & 64.83  & 58.57\\\hline
\end{tabular}%
\end{adjustbox}
\caption{Experiments using linear classifier with features based on word embeddings. Trained on \textit{Dev} (tuned with cross-validation) and evaluated on \textit{Test}.}
\label{table:feats-embeddings}
\end{table*}

We experiment with different word embedding models pre-trained on a large number of tokens including \textit{word2vec}\footnote{https://code.google.com/archive/p/word2vec/ - Pre-trained embeddings on Google News dataset (100B words).} \cite{mikolov-yih-zweig:2013:NAACL-HLT}, \textit{GloVe} \cite{Pennington2014} and \textit{ConceptNet Numberbatch} \cite{speer2016ensemble}.
Results on training the feature-based model with different word embeddings are shown in Table \ref{table:feats-embeddings}. The results indicate how well the vector representation models perform in terms of encoding common sense stories. We present the performance of the embedding models depending on the defined features. We perform feature ablation experiments to determine the features which contribute most to the overall score for the different models. Using {\em All} features defined in Section \ref{sec:method-baseline:features}, the word2vec vectors, trained on Google News 100B corpus perform best followed by ConcepNet enriched embeddings and Glove trained on Common Crawl 840B. The word2vec model suffers most when similarity features are excluded. We note that the ConceptNet embeddings do not decrease performance when similarity features are excluded, unlike all other models. 
We also see that the \textit{POS similarities} are more important than the \textit{MaxSim} and the \textit{Sim (cosine betwen all words in \textit{Story} and \textit{Ending})} as they yield worse results, for almost all models, when excluded from \textit{All} features.

In column \textit{WE E1, E2} we report results on features based only on {\em Ending1} and {\em Ending 2}. We note that the overall results are still very good. From this we can derive that the difference of \textit{Good vs.\ Bad}  endings is not only defined in the story context but it is also characterized by the information present in these sentences in isolation.
This could be due to a reporting bias \cite{Gordon2013Commonsense-Bias} employed by the crowd-workers in the corpus construction process. 

The last column \textit{Sims only} shows results with features based only on similarity features. It includes all story-to-ending semantic vector similarities described in Section \ref{sec:method-baseline:features}. 

We also perform experiments with the neural LSTM model. In Table  \ref{table:neural:comparison} we compare results of the LSTM representation models that we examined for the task. We trained the models on the \textit{Dev-Train} for 10 epochs and take the best performing model on the \textit{Dev-Dev} dataset.

\begin{table}[]
\centering
\begin{tabular}{llll}
\textbf{Model} & \textbf{Epoch} & \textbf{Dev-Dev} & \textbf{Test} \\\hline
LSTM Raw             & 7     & 77.12    & 72.10 \\
LSTM Raw + Att & 2     & 79.25    & 68.30 \\
Attention            & 9     & 72.79    & 63.22\\\hline
\end{tabular}
\caption{Comparison between LSTM representation strategies.}
\label{table:neural:comparison}
\end{table}

Our best LSTM model uses only raw LSTM encodings of the \textit{Story} and the candidate \textit{Endings}, without using attention. Here the \textit{Attention} representation is intended to capture semantic relations between the \textit{Story} context and the candidate \textit{Endings}, similar to the \textit{Similarities only} setup examined with the feature-based approach.
Considering the low performance of {\em Attention},
the poor 
results of the \textit{semantic similarity features} and the high performance of the feature-based model with {\em Ending only} features we hypothesize that the reason for this unexpected result is that the background knowledge presented in the training data is not enough to learn strong relations between the story context and the endings. 

\paragraph{Experiments with generated data.} We also try to employ the data from the ROC Stories corpus by generating labeled datasets following all approaches described in Section \ref{sec:data}. Training our best neural model using all of the generated datasets separately without any further data selection yields results close to the random baseline of the ending selection task. We also try to filter the generated data by training several feature-based and neural models
with our best configurations and evaluating the generated data. We take only instances that have been classified correctly by all models. The idea here was to generate much more data (with richer vocabulary) that performs at least as good as the \textit{Dev} data as training. However the results of the models trained on these datasets were not better than the one trained on \textit{Dev} and \textit{Dev-Dev (for the neural models)}.

\section{Conclusion and Future work}
\label{sec:future}
In this work we built two strong supervised baseline systems for the Story Cloze task: one based on semantic features based on word embedding representations and bi-clausal similarity features obtained from them, and one on based on a neural network LSTM-based encoder model. The neural network approach trained on a small dataset performs worse than the feature-based classifier by a small margin only, and our best model ranks 3rd according to the shared task web page. 

In terms of data, it seems that the most important features are coming from word representations trained on large text corpora rather than relations between the data. Also we can train a model that performs well only on the given endings, without a given context which could mean that there is a bias in the annotation process. However, this requires more insights and analysis.  

In future work we plan improve the current results on this (or a revised) dataset by collecting more external knowledge and obtaining more or different training data from the original ROC Stories corpus.

\textbf{Acknowledgments.} This work is supported by the German Research Foundation as part of the Research Training Group ``Adaptive Preparation of Information from Heterogeneous Sources'' (AIPHES) under grant No. GRK 1994/1. 

\bibliography{bib}
\bibliographystyle{eacl2017}

\end{document}